\documentclass[letterpaper]{article} 
\usepackage{aaai25}
\usepackage{times}  
\usepackage{helvet}  
\usepackage{courier}  
\usepackage[hyphens]{url}  
\usepackage{graphicx} 
\usepackage{natbib}  
\usepackage{caption} 
\usepackage{algorithm}
\usepackage{algorithmic}

\usepackage{newfloat}
\usepackage{listings}
\DeclareCaptionStyle{ruled}{labelfont=normalfont,labelsep=colon,strut=off} 
\floatstyle{ruled}
\newfloat{listing}{tb}{lst}{}
\floatname{listing}{Listing}
\usepackage{amsmath,amsfonts,amssymb}
\usepackage{multirow}
\usepackage[justification=centering]{caption}

\usepackage{amsthm}
\newtheorem{theorem}{Theorem}

\title{Mining Unstructured Medical Texts With Conformal Active Learning}
\author {
    Juliano Genari\textsuperscript{\rm 1},
    Guilherme Tegoni Goedert\textsuperscript{\rm 1}
}
\affiliations {
    \textsuperscript{\rm 1}Escola de Matemática Aplicada, Fundação Getúlio Vargas, Rio de Janeiro, Rio de Janeiro, Brazil\\
    araujo.juliano@fgv.edu.br, guilherme.goedert@fgv.br
}

\usepackage{bibentry}

\begin{document}

\maketitle

\begin{abstract}
The extraction of relevant data from Electronic Health Records (EHRs) is crucial to identifying symptoms and automating epidemiological surveillance processes. By harnessing the vast amount of unstructured text in EHRs, we can detect patterns that indicate the onset of disease outbreaks, enabling faster, more targeted public health responses. Our proposed framework provides a flexible and efficient solution for mining data from unstructured texts, significantly reducing the need for extensive manual labeling by specialists. Experiments show that our framework achieving strong performance with as few as 200 manually labeled texts, even for complex classification problems. Additionally, our approach can function with simple lightweight models, achieving competitive and occasionally even better results compared to more resource-intensive deep learning models. This capability not only accelerates processing times but also preserves patient privacy, as the data can be processed on weaker on-site hardware rather than being transferred to external systems. Our methodology, therefore, offers a practical, scalable, and privacy-conscious approach to real-time epidemiological monitoring, equipping health institutions to respond rapidly and effectively to emerging health threats.
\end{abstract}

%

\section{Introduction}

Epidemiological surveillance, the practice of tracking health indicators to identify potential outbreaks, is a critical yet often neglected task in many healthcare systems. Despite its importance, epidemiological surveillance remains costly and labor-intensive, requiring significant time and expertise to extract insights from extensive health data \cite{thacker1988method, ibrahim2020epidemiologic}. The challenge is particularly pronounced with Electronic Health Records (EHRs), where valuable information about symptoms and emerging patterns is often stored in unstructured text formats that can include unclear or inconsistent language. Many institutions lack the necessary staffing to routinely mine and analyze this data, leading to missed opportunities for early intervention and epidemic prevention \cite{kelly2019importance}. Addressing this gap requires a scalable, efficient solution that reduces the reliance on specialist input while ensuring the quality and accuracy of the insights produced.

We propose a framework that introduces an innovative approach to automate and streamline the surveillance process. By combining active learning methods (cf., the survey by \citealt{settles2009active}) with label-conditional conformal prediction \cite{vovk2012conditional}, our methodology leverages machine learning to identify relevant information within unstructured text and improve predictive performance without exhaustive manual labeling. This framework allows us to make the most of any classification model capable of producing text embeddings, enabling it to adapt to the computational constraints of various deployment environments. The active learning component prioritizes the most informative data points for specialist review, while periodically checking points the model predicted with high certainty, significantly reducing the labeling burden while ensuring that key data is accurately categorized.

Furthermore, our framework provides reliability by leveraging conformal prediction, which ensures that the predicted labels fall within a specified confidence level $\alpha$ \cite{vovk2012conditional}. This guarantee offers healthcare providers a transparent and interpretable uncertainty-aware guarantee of correctness for each classification, enabling them to make well-informed decisions based on dependable insights. By utilizing simpler, adaptable models that can be deployed directly within healthcare institutions, our framework also addresses critical privacy concerns, as sensitive patient data remains on-site without requiring transfer to external systems. Ultimately, our approach empowers healthcare facilities to perform timely and accurate epidemiological surveillance, transforming EHR data into actionable insights while optimizing resource usage and protecting patient privacy.

Our main contributions are: (1) a novel Conformal Active Learning framework that combines active learning with label-conditional conformal prediction, offering reliable predictions while minimizing manual labeling; (2) a model-agnostic design that works with any classification model capable of generating embeddings, allowing flexibility for deployment in hardware-constrained healthcare facilities; (3) a clustering-based selection process that improves performance by ensuring diversity on the texts selected for manual labelling; and (4) the release of open-source code under GPL-v3.0 and a user-friendly web interface, OLIM to facilitate deployment and accessibility. These contributions address critical challenges in unstructured data processing, privacy-preserving deployment, and iterative machine learning for healthcare.

\section{Conformal Active Learning}

\begin{figure}[t]
\centering
\includegraphics[width=1\columnwidth]{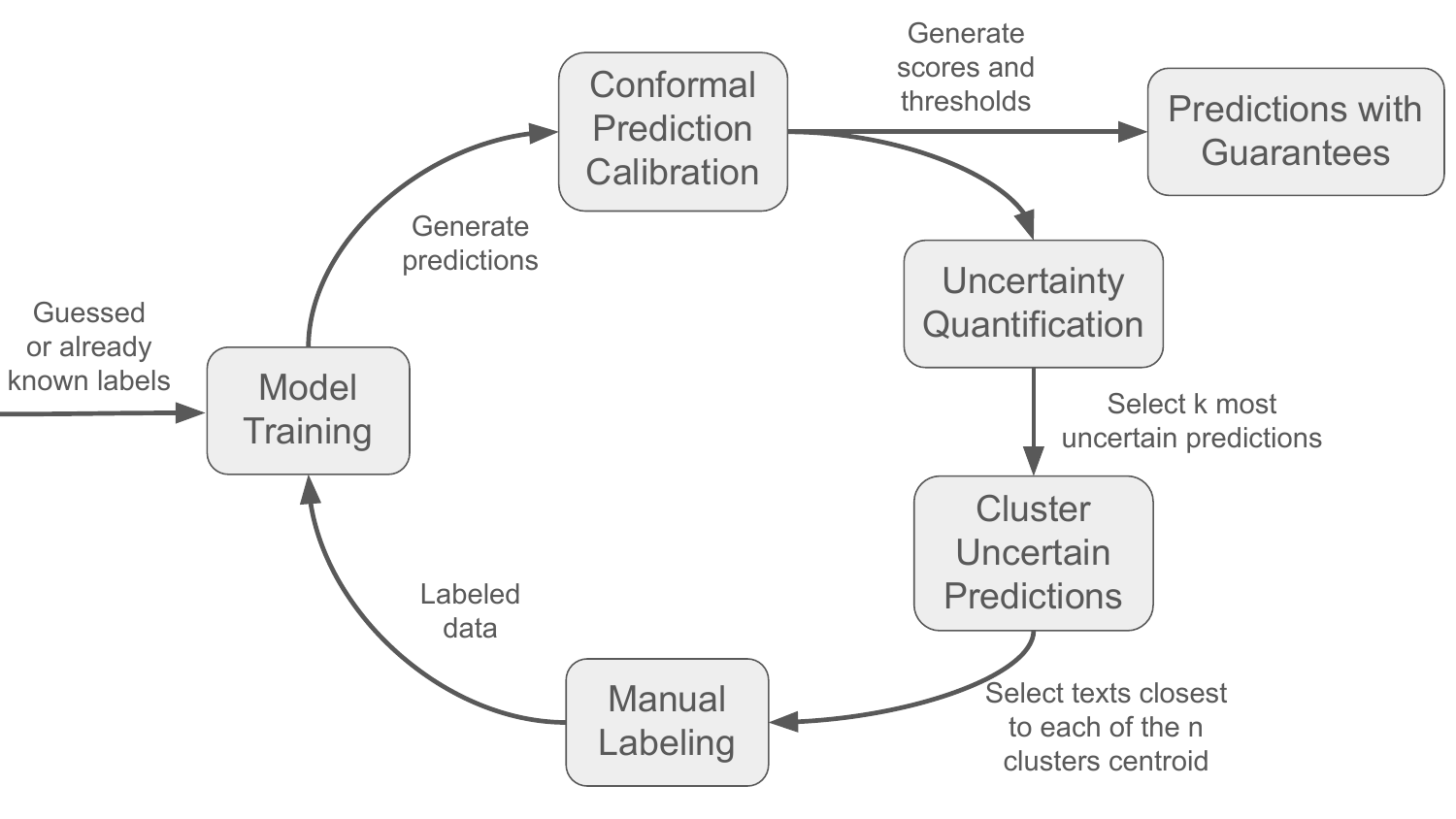}
\caption{Diagram of the active leaning cycle.}
\label{fig:active}
\end{figure}

Our main goal is to infer labels $Y$ (for example, if a patient has or not a symptom) for unstructured texts $X$ (EHRs, etc.). To achieve this goal we propose a Conformal Active Learning framework, see Figure \ref{fig:active}, designed to reduce the amount of manual labeling required to achieve high-quality predictions. 

This process begins with either an initial set of guessed labels (generated by an uninformed model, using a zero-shot classification model or simply randomly assigning) or a small subset of manually labeled data, which we use to train a preliminary classification model. Using this initial model, we can start generating predictions for the remaining unlabeled data, refining the model iteratively through cycles of selective labeling and retraining.

\subsection{Conformal Scores}

All data inputted on the active learning pipeline is split on training and validation datasets. On each cycle a new classification model is trained, after which we need to quantify the uncertainty of each prediction. To achieve such task we first calibrate a label-conditional conformal model on the validation set. This model allows each point $X_i$ to be associated with a conformal score, indicating how uncertain the model is about a given label $y$, by the means of a \emph{conformity score function}, in this case deriving directly from the model's logits\footnote{There are many possible conformity scores for classification -- e.g., \cite{conformityscore-classification-1,conformityscore-classification-2}. The freedom in the choice of conformity scores is a great feature of our approach, allowing us to leverage insights from the conformal prediction literature for better uncertainty quantification.}:
\begin{equation}
    s(x, y) = 1 - \hat{p}(y \ |\ x),
    \label{eq:scorey}
\end{equation}
where $\hat{p}(y | x)$ is the predicted probability of $x$ to have the label $y$ according to the classification model. To calibrate the label-conditional conformal model we compute a thresholds for each label $y$ by leveraging the data in the validation split, $(X_1, Y_1), \ldots, (X_n, Y_n)$:
\begin{equation}
    t_{y,\alpha} = \hat{q}_{1-\alpha} (\{s(X_i,Y_i) : Y_i = y \} \cup \{+\infty\}),
\end{equation}
where $\hat{q}_{1-\alpha}$ is the $1-\alpha$ empirical quantile. With this we can make calibrated set predictions for new $X^\mathrm{new}$s, consisting of a subset of labels for $X^\mathrm{new}$:
\begin{equation}
    C_\alpha (X^\mathrm{new}) = \{ y : s(X^\mathrm{new}, y) \leq t_{y,\alpha} \}.
\end{equation}
This construction assures us of the strong guarantees of label-conditional conformal prediction:
\begin{theorem}
    Suppose that the data $(X_1, Y_1), \ldots, (X_n, Y_n)$, along with a new data point $(X^\mathrm{new}, Y^\mathrm{new})$, are exchangeable random variables. Then,
    for any conformity score $s$, any $\alpha \in (0, 1)$, it holds that, for all label values $y$,
    \[ \mathbb{P}[ Y^\mathrm{new} \in C_\alpha (X^\mathrm{new}) \ |\ Y^\mathrm{new} = y ] \geq 1 - \alpha. \]
\end{theorem}

\subsection{Ranking Samples by Uncertainty}

To score each prediction we take the mean conformity score over all the labels $y$ predicted for a sample text $X$:
\begin{equation}
    S_{X} = \frac{1}{|C_\alpha(X)|} \sum_{y \in C_\alpha(X)} s(X, y)
\end{equation}
These scores serve as a basis for ranking data points where the model’s predictions are less certain, allowing us to prioritize them in the selection process.

\subsection{Clustering Selection for Manual Labeling}

After calculating conformal scores, we rank unlabeled data points by uncertainty and select the top $k_{top}$ with the highest scores. To ensure diversity, we use $k$-means to cluster these high uncertainty points into $k_{cluster}$ ($k_{cluster}<k_{top}$) groups. This step relies on the classification model’s embeddings, which enable $k$-means to form meaningful clusters. From each cluster, the point closest to the centroid is selected, yielding a diverse sample of $k_{cluster}$ data points for manual labeling that represents a broad range of high uncertainty predictions.

Clustering avoids redundancy and ensures labeled data is as informative as possible. Additionally, our framework provides flexibility during selection. While the model retrains, the $k_{cluster}$ selected points can be labeled. If these are used up before retraining completes, the framework randomly selects from the remaining $k_{top}$ high uncertainty points to ensure uninterrupted progress.

The framework can also include a fraction of low uncertainty predictions in the $k_{top}$ points before the clustering step. This continually validates the model's performance on straightforward cases, further enhancing reliability. Incorporating both high and low uncertainty points during labeling has shown good results, improving both performance and confidence in predictions.

\section{Deployability}

The proposed framework is classification model agnostic, requiring only that the model generates embeddings for data points. This flexibility allows for model training and selection based on hardware available on-premise in healthcare institutions, ensuring the framework can adapt to a wide range of deployment environments. On-premise deployment capabilities also address critical privacy concerns, as sensitive patient data remains securely within the institution rather than being transferred to external systems. This allows external access to aggregated structured data, permitting government and research institutions to generate powerful data-driven insights for epidemiological surveillance while protecting patient information.

To further support ease of use, we are developing an accompanying web interface (see Appendix C), OLIM (Open Labeller for Iterative Machine learning), to streamline the iterative labeling process and enhance user experience. All code for our framework\footnote{\url{https://gitlab.com/nanogennari/olim-learner/-/tree/DAI-2025}} and OLIM \footnote{\url{https://gitlab.com/nanogennari/olim}} is released as open source under the GPL-v3.0 license, ensuring transparency and community-driven improvements. For seamless deployment, the entire setup is packaged in Docker containers, allowing for quick and straightforward installation across diverse environments. This design makes our framework highly accessible and adaptable, empowering healthcare institutions and government entities to perform iterative machine learning on-site with minimal setup.

\section{Experimental Setup}

\paragraph{Dataset} Given the sensitive nature of patient information, we will demonstrate our proposed framework on a database of Amazon product reviews \cite{amazonReviews}, which presents challenges similar to those found in medical records. Product reviews often feature unclear or nonstandard language to describe subjective issues such as product quality, defects, and consumer satisfaction. We can also infer the type of product from the text as a task similar to symptom identification. This dataset serves as a suitable proxy, allowing us to test our framework's ability to process unstructured and diverse language effectively.

Notably, there are no publicly available open medical databases for this purpose. While databases such as MIMIC-III \cite{johnson2016mimic} exist, they do not contain electronic health record (EHR) data due to the high sensitivity of patient information and the significant challenges involved in anonymization of unstructured data. By using this proxy dataset, we refine the active learning and conformal prediction components, preparing our framework for deployment in real-world healthcare scenarios while addressing comparable linguistic complexities.

\paragraph{Labels} We defined four labels for classification: \textit{Pet product}, \textit{Drinkable product}, \textit{Low quality product}, and \textit{Damaged product}. The first two labels represent common categories, while the third is highly subjective, and the fourth is a rarer label. This variety of labels was chosen to challenge our framework in scenarios ranging from frequent, well-defined classes to sparse or ambiguous cases. 

\paragraph{Parameters} All experiments were conducted with 100 or 200 manually labeled texts. We used a $k_{top} = 500$ and $k_{cluster} = 6$. The initial ranking was done randomly and the first training cycle was done after $10$ labels where inputted. Additionally, $20\%$ of the labeled data was randomly assigned to the validation dataset in each cycle to calibrate the conformal prediction component and ensure reliable uncertainty estimation. All evaluations were done with a confidence level of $90\%$.

\paragraph{Models} Our framework is model agnostic and can work with any classification model capable of generating embeddings. We tested it across different classification models to demonstrate its flexibility. The first model was \textbf{TF-IDF + XGBClassifier} \cite{chen2016xgboost}, a lightweight text classification model with good performance suitable for scenarios where computational resources are very limited. The second was \textbf{DeBERTaV3} \cite{he2020deberta} \footnote{We used the available \texttt{deberta\_v3\_base\_en} pre-trained weights.}, a more advanced deep learning transformer-based model that simulates settings with moderate computational resources but is still feasible for on-premise deployment in healthcare facilities with commodity hardware. 

\paragraph{Additional tests} For the \textit{Pet product} label, we additionally experimented with a 50/50 and a 70/30 split of high and low uncertainty texts within the selected $k_{top}$ points to evaluate the impact of balancing exploration and validation during manual labeling. For the \textit{Damaged product} label, we tested the number of manually labeled texts with 100 and 200 texts to analyze how increased supervision affects our framework's performance. Furthermore, we tested starting with 40 pre-labeled texts to assess the impact of initial labeled data. In this case, positive samples were identified using manual keyword searches, while negatives were selected from a random list. And lastly we tested labeling a random list of 100 texts of the labels \textit{Pet product}, to represent common labels, and \textit{Damaged product}, to represent rare labels. A \textbf{TF-IDF + XGBClassifier} model was trained on the randomly selected labeled data to compare with data selected by our framework.

\section{Results}

\bgroup
\def\arraystretch{1.2}
\begin{table}[t]
\centering
\begin{tabular}{lccc}
\textbf{Label}      & \textbf{Accuracy} & \textbf{AUC-ROC} & \textbf{Yes/No} \\ \hline
Pet prod.           & 0.92 $\pm$ 0.01   & 0.94 $\pm$ 0.06    & 62/138          \\ \hline
Drinkable           & 0.85 $\pm$ 0.01   & 0.82 $\pm$ 0.04    & 70/130          \\ \hline
Low quality         & 0.77 $\pm$ 0.01   & 0.79 $\pm$ 0.04    & 46/154          \\ \hline
Damaged$^1$         & 0.91 $\pm$ 0.01   & 0.75 $\pm$ 0.08    & 39/161          \\ \hline
\end{tabular}
\caption{Final performance with \textbf{XGBClassifier} for the proposed labels after 200 manual labels using our framework, with $k_{top}$ splitted 30/70 on high and low uncertainty, started with 20 pre-labelled texts. \\ $^1$Started with 40 pre-labelled texts.}
\label{tab:metrics_final}
\end{table}
\egroup

We achieved good results even with a simple model such as \textbf{TF-IDF + XGBClassifier}. With 100 manual labels, the performance, as shown in Table \ref{tab:metrics}, still leaves room for improvement in some cases, particularly for less frequent labels. However, when combining high and low uncertainty in $k_{top}$ selection and adding a small number of pre-labeled texts (20), we managed to achieve strong results with just 200 manual labels; see Table \ref{tab:metrics_final}. This represents a remarkably low number of labeled texts for a complex classification problem, demonstrating the effectiveness of the proposed framework in optimizing labeling efforts while maintaining robust performance. A more complete table of results, including additional experimental configurations and metrics, can be found in Appendix B.

\paragraph{Deep Learning Models} Our results with deep learning models, such as \textbf{DeBERTaV3}, were not as strong as initially expected, see Table \ref{tab:metrics_NN}. Surprisingly, the lightweight \textbf{TF-IDF + XGBClassifier} demonstrated competitive performance, underscoring the potential of simpler models to deliver good results under resource-constrained environments. While deep learning models hold promise for improving classification performance, particularly in zero-shot and few-shot learning scenarios for bootstrapping the framework, further exploration is needed to identify more performant models. One potential direction is a hybrid approach: leveraging a deep learning model for the initial bootstrapping phase to generate pre-labeled examples, then employing a simpler model to guide uncertainty-based active learning cycles. Finally, a more complex deep learning model could be deployed for the final predictions once sufficient labeled data is available.

\bgroup
\def\arraystretch{1.2}
\begin{table}[t]
\centering
\begin{tabular}{lccc}
\textbf{Label}      & \textbf{Accuracy} & \textbf{AUC-ROC} & \textbf{Yes/No} \\ \hline
Pet prod.           & 0.44 $\pm$ 0.06   & 0.66 $\pm$ 0.02    &  23/77          \\ \hline
Drinkable           & 0.74 $\pm$ 0.02   & 0.42 $\pm$ 0.08    &  32/67          \\ \hline
\end{tabular}
\caption{Final performance with \textbf{DeBERTaV3} after 100 manual labels using our framework.}
\label{tab:metrics_NN}
\end{table}
\egroup

\begin{figure}[t]
    \centering
    \includegraphics[width=0.95\linewidth]{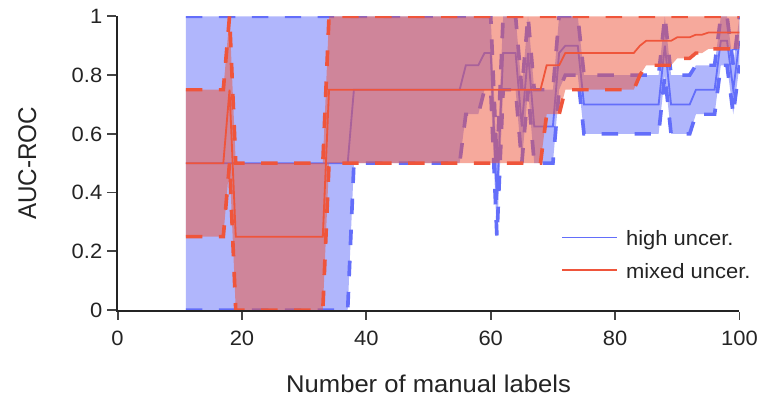}
    \caption{Convergence of AUC-ROC for the \textit{Pet product} label, using the \textbf{XGBBoost} model, with only high uncertainty and a 70/30 mix of high and low uncertainty.}
    \label{fig:convergence}
\end{figure}

\paragraph{Mix of high and low uncertainty} Incorporating a mix of high and low uncertainty texts during the selection process significantly improves the framework's performance, as observed in the \textit{Pet product} label. By using a 50/50 or a 70/30 split of high and low uncertainty points, the AUC-ROC score increased from $0.79$ (using only high uncertainty predictions) to $0.95$, demonstrating a substantial boost in classification performance. On Figure \ref{fig:convergence} we can also see a more stable convergence of the AUC-ROC with mixed high and low uncertainty. Additionally, this approach improved other metrics such as accuracy and precision, which are more meaningful in this case due to the less unbalanced nature of the select texts. The 70/30 split also managed to include 36 positive samples for labeling, compared to only 23 positives when exclusively focusing on high uncertainty predictions. This result highlights the importance of combining exploratory labeling of uncertain points with validation of straightforward cases to maximize the diversity and informativeness of the labeled dataset.

\bgroup
\def\arraystretch{1.2}
\begin{table}[t]
\centering
\begin{tabular}{lccc}
\textbf{Label}      & \textbf{Accuracy} & \textbf{AUC-ROC} & \textbf{Yes/No} \\ \hline
Pet prod.           & 0.83 $\pm$ 0.02   & 0.79 $\pm$ 0.07    & 23/77           \\ \hline
Pet prod.$^1$       & 0.80 $\pm$ 0.02   & 0.95 $\pm$ 0.05    & 27/73           \\ \hline
Damaged             & 0.97 $\pm$ 0.03   & 0.50 $\pm$ 0.50    & 2/98            \\ \hline
Damaged$^2$         & 0.96 $\pm$ 0.01   & 0.75 $\pm$ 0.25    & 5/195           \\ \hline
Damaged$^3$         & 0.82 $\pm$ 0.04   & 0.88 $\pm$ 0.12    & 30/70           \\ \hline
\end{tabular}
\caption{Final performance with \textbf{XGBClassifier} after 100 manual labels using our framework. \\ {\small $^1$Done with $k_{top}$ splitted 50/50 on high and low uncertainty. \\ $^2$Done with 200 manual labelings. \\ $^3$Started with 40 pre-labelled texts.}}
\label{tab:metrics}
\end{table}
\egroup

\paragraph{Less frequent labels} For less frequent labels, such as \textit{Damaged product}, the framework can require a significant number of manual labels to achieve meaningful convergence. As shown in Table \ref{tab:metrics}, even with 200 manually labeled texts, the framework struggled to identify many positive instances, resulting in poor performance reflected by an AUC-ROC score of $0.5$. This highlights the challenge of working with sparse or rare labels, where the limited representation of positive examples in the dataset hinders the model's ability to generalize effectively. However, when provided with pre-labeled texts to initialize the framework, performance improves substantially. when we start including 40 pre-labeled examples, with only 100 manually labeled texts the framework achieved a significantly higher AUC-ROC score of $0.88$. Furthermore, it successfully selected an additional 10 positive texts on top of the 20 positive examples it started with, demonstrating that a small initial labeled set can dramatically enhance our framework's ability to handle less frequent labels. This limitation can also be easily addressed by incorporating automated techniques to generate an initial set of texts to be manually labeled. For example, a simple keyword search could identify a preliminary set of positive examples, or a zero-shot classifier could be used to assign initial labels based on textual descriptions of the target category.

\paragraph{Randomly selecting data} on more common labels, such as \textit{Pet product}, can achieve results close to our framework (see Appendix B). In our test, we found 18 positive cases out of 100 and trained a \textbf{TF-IDF + XGBClassifier} model with an AUC-ROC of $0.90 \pm 0.10$. But when tested with more rare labels the performance of models trained on randomly select data quickly degrades compared to our active learning framework, on the label \textit{Damaged product} random selection could only find one positive case out of 200 texts, resulting in an unusable model with AUC-ROC of $0.50 \pm 0.50$, which compared to Table \ref{tab:metrics} highlights the far superior performance of our active learning framework in identifying relevant entries.

\section{Conclusion}

In this work, we proposed a novel Conformal Active Learning framework designed to reduce manual labeling efforts while ensuring reliable and interpretable predictions. By integrating active learning with label-conditional conformal prediction, our framework enables efficient handling of unstructured text data. 

Through extensive experiments on a proxy dataset of Amazon reviews, results show that we can achieve good performance with a low number of specialist-labeled texts. Significantly optimizing the labeling process. Incorporating a mix of high and low uncertainty points and leveraging pre-labeled examples further enhances performance. While deep learning models hold promise for further improvements, particularly in bootstrapping the framework.

By making our framework open source and accessible, with deployment tools like OLIM, we aim to empower healthcare institutions to perform privacy-preserving, real-time epidemiological surveillance. Future work will involve testing on real medical data and further refining the framework for practical deployment in healthcare settings.

\nocite{*}
\bibliography{bibliography}

\end{document}


\maketitle

\appendix

\section{Proof of Theorem 1}

\begin{theorem}
    Suppose that the data $(X_1, Y_1), \ldots, (X_n, Y_n)$, along with a new data point $(X^\mathrm{new}, Y^\mathrm{new})$, are exchangeable random variables. Then,
    for any conformity score $s$, any $\alpha \in (0, 1)$, it holds that, for all label values $y$,
    \[ \mathbb{P}[ Y^\mathrm{new} \in C_\alpha (X^\mathrm{new}) \ |\ Y^\mathrm{new} = y ] \geq 1 - \alpha. \]
\end{theorem}

\begin{proof}
    Since the data is exchangeable, the random variables $S_i = s(X_i, Y_i)$ for $i = 1, \ldots, n, \mathrm{new}$ are also exchangeable. Thus, for any $y$:
    \begin{align*}
        \mathbb{P}[ Y^\mathrm{new} \in C_\alpha (X^\mathrm{new}) \ |\ Y^\mathrm{new} = y ]
        &= \mathbb{P}[ s(X^\mathrm{new}, Y^\mathrm{new}) \leq t_{y,\alpha} \ |\ Y^\mathrm{new} = y]
        \\ &= \mathbb{P}[ S_\mathrm{new} \leq \hat{q}_{1-\alpha} (\{S_i : i = 1, \ldots, n \text{ such that } Y_i = y \} \cup \{+\infty\}) \ |\ Y^\mathrm{new} = y]
        \\ &\geq 1 - \alpha,
    \end{align*}
    where the last step holds by Lemma 1 of (Tibshirani et al. 2019). 
\end{proof}

\newpage
\section{Comprehensive Results Overview}

Table \ref{tab:comp} presents the final performance of our framework using different classification models and across various configurations. The results highlight differences in accuracy, precision, recall and AUC-ROC scores, demonstrating the adaptability of the framework.

\setcounter{table}{3}
\bgroup
\def\arraystretch{1.3}
\begin{table}[h]
\begin{tabular}{lccccccl}
Label                        & Model      & Accuracy        & Precision       & Recall          & AUC-ROC       & Yes/No & Obs.     \\ \hline
\multirow{6}{*}{Pet}     & TF-IDF+XGB & 0.83 $\pm$ 0.02 & 0.81 $\pm$ 0.03 & 0.97 $\pm$ 0.03 & 0.79 $\pm$ 0.07 & 23/77  &          \\ \cline{2-8} 
                             & TF-IDF+XGB & 0.80 $\pm$ 0.02 & 0.74 $\pm$ 0.03 & 0.96 $\pm$ 0.04 & 0.95 $\pm$ 0.05 & 27/73  &  $^{1}$        \\ \cline{2-8} 
                             & TF-IDF+XGB & 0.84 $\pm$ 0.03 & 0.84 $\pm$ 0.03 & 0.96 $\pm$ 0.04 & 0.95 $\pm$ 0.05 & 36/64  &  $^{2}$        \\ \cline{2-8} 
                             & TF-IDF+XGB & 0.81 $\pm$ 0.02 & 0.84 $\pm$ 0.03 & 0.90 $\pm$ 0.03 & 0.85 $\pm$ 0.05 & 41/59  &  $^{3}$        \\ \cline{2-8} 
                             & TF-IDF+XGB & 0.92 $\pm$ 0.01 & 0.96 $\pm$ 0.01 & 0.96 $\pm$ 0.01 & 0.94 $\pm$ 0.06 & 70/130 &  $^{245}$        \\ \cline{2-8} 
                             & DeBERTaV3  & 0.66 $\pm$ 0.02 & 0.66 $\pm$ 0.02 & 0.97 $\pm$ 0.03 & 0.44 $\pm$ 0.06 & 23/77  &          \\ \cline{2-8} 
                             & TF-IDF+XGB+Rand & 0.84 $\pm$ 0.02 & 0.83 $\pm$ 0.02 & 0.97 $\pm$ 0.03 & 0.90 $\pm$ 0.10 & 18/82  & $^{7}$         \\ \hline
\multirow{3}{*}{Drinkable}   & TF-IDF+XGB & 0.98 $\pm$ 0.02 & 0.98 $\pm$ 0.02 & 0.98 $\pm$ 0.02 & 0.50 $\pm$ 0.50 & 19/81  &          \\ \cline{2-8} 
                             & TF-IDF+XGB & 0.85 $\pm$ 0.01 & 0.86 $\pm$ 0.02 & 0.92 $\pm$ 0.02 & 0.82 $\pm$ 0.04 & 70/130 &  $^{245}$        \\ \cline{2-8} 
                             & DeBERTaV3  & 0.74 $\pm$ 0.02 & 0.74 $\pm$ 0.02 & 0.97 $\pm$ 0.03 & 0.42 $\pm$ 0.08 & 32/67  &          \\ \hline
\multirow{2}{*}{Low quality} & TF-IDF+XGB & 0.74 $\pm$ 0.02 & 0.78 $\pm$ 0.03 & 0.91 $\pm$ 0.03 & 0.50 $\pm$ 0.10 & 20/80  &          \\ \cline{2-8} 
                             & TF-IDF+XGB & 0.77 $\pm$ 0.01 & 0.76 $\pm$ 0.01 & 0.98 $\pm$ 0.02 & 0.79 $\pm$ 0.04 & 46/154 & $^{245}$         \\ \hline 
\multirow{4}{*}{Damaged}      & TF-IDF+XGB & -- & -- & -- & 0.50 $\pm$ 0.50 & 2/98   &          \\ \cline{2-8} 
                             & TF-IDF+XGB & 0.96 $\pm$ 0.01 & 0.96 $\pm$ 0.01 & 0.99 $\pm$ 0.01 & 0.75 $\pm$ 0.25 & 5/195  &  $^{4}$        \\ \cline{2-8} 
                             & TF-IDF+XGB & 0.82 $\pm$ 0.04 & 0.81 $\pm$ 0.04 & 0.95 $\pm$ 0.05 & 0.88 $\pm$ 0.12 & 30/70  &  $^{6}$        \\ \cline{2-8} 
                             & TF-IDF+XGB & 0.91 $\pm$ 0.01 & 0.91 $\pm$ 0.01 & 0.99 $\pm$ 0.01 & 0.75 $\pm$ 0.08 & 39/161 &  $^{245}$        \\ \cline{2-8} 
                             & TF-IDF+XGB+Rand & -- & -- & -- & 0.50 $\pm$ 0.50 & 1/199  & $^{47}$         \\  \hline 
\end{tabular}
\caption{Final performance with different classification models for the proposed labels after 100 manual labels using our framework. \\ {\small $^1$Done with $k$ splitted 50/50 on high and low uncertainty. \\ $^2$Done with $k$ splitted 30/70 on high and low uncertainty. \\ $^3$Done with $k$ splitted 70/30 on high and low uncertainty. \\ $^4$Done with 200 manual labelings. \\ $^5$Started with 20 pre-labelled texts.\\ $^6$Started with 40 pre-labelled texts.\\ $^7$ Random selection of entries no active learning.}}
\label{tab:comp}
\end{table}
\egroup

\newpage
\section{The OLIM User Interface}

OLIM (Open Labeller for Interactive Machine Learning) is a web interface designed to streamline dataset annotation while integrating  your conformal active learning framework. Its Dockerized architecture ensures scalability, and its intuitive design caters to both technical and non-technical users.

\begin{figure}[ht]
    \centering
    \includegraphics[width=0.75\textwidth]{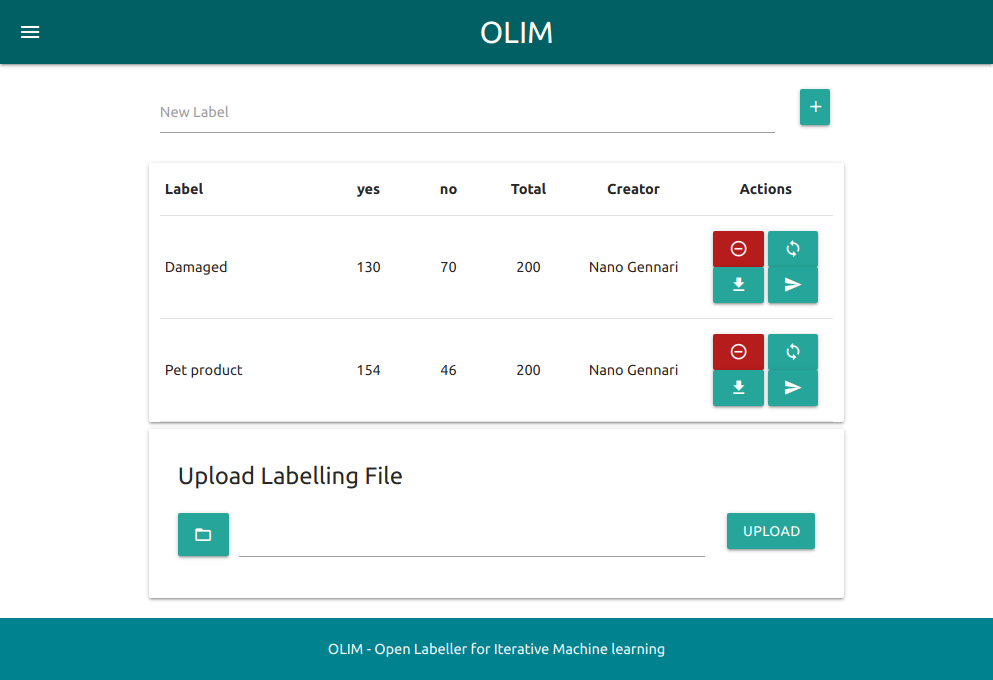}
    \caption{Label management interface with label management, progress, CSV download and upload and option to enter individual Active Learning.}
    \label{fig:label_management}
\end{figure}

\paragraph{User and Label Management}  
The interface provides role-based access control, enabling user type specific access. Users can create, delete, or export labels as CSV files for offline analysis, while real-time metrics display labeling progress. As shown in Figure~\ref{fig:label_management}, the dashboard centralizes these tasks, including bulk operations and visualization of label distribution.

\paragraph{Active Learning Integration}  
The OLIM interface directly integrates with \textbf{OLIM-Learner}, an implementation of our proposed active learning framework. Users can launch the active learning pipeline directly from the labels management page (Figure~\ref{fig:label_management}) or manually sync labels created outside the framework. Labeling queues dynamically prioritize uncertain or contentious samples identified by the framework, streamlining the annotation workflow. For flexibility, labels modified externally (e.g., bulk CSV uploads) can be synced to OLIM-Learner to trigger retraining.

\begin{figure}[ht]
    \centering
    \includegraphics[width=0.75\textwidth]{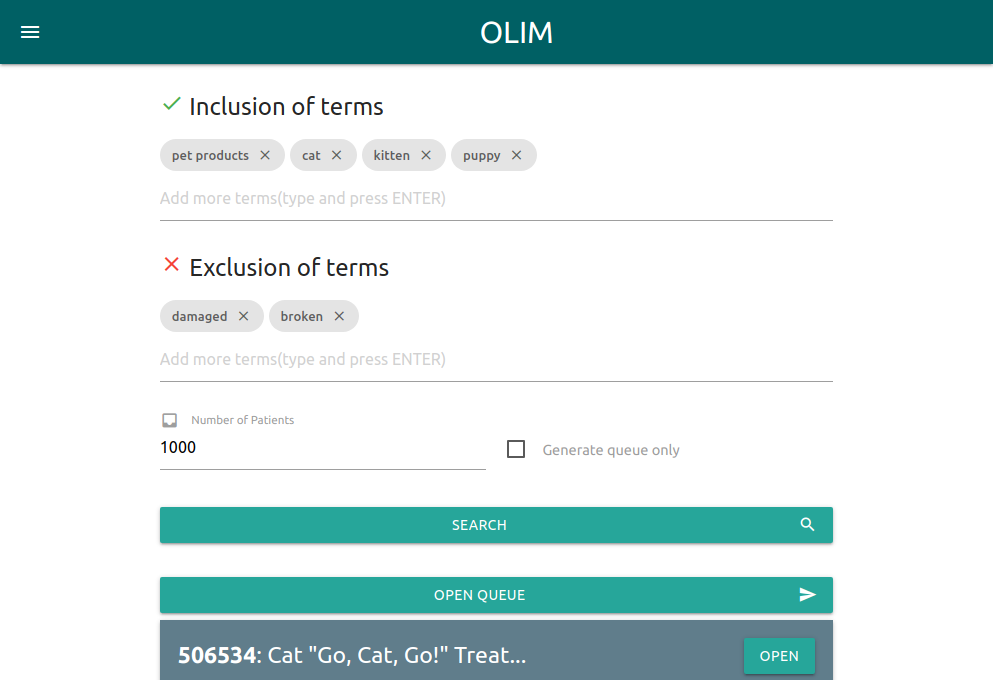}
    \caption{Search interface with term inclusion/exclusion syntax.}
    \label{fig:search_interface}
\end{figure}

\paragraph{Search Interface}  
Powered by Elasticsearch, the search engine supports inclusion/exclusion of terms. Figure~\ref{fig:search_interface} demonstrates the query panel, which highlights matching terms in results and provides histograms of label frequencies. Users can save frequently used filters, such as date ranges or user-specific annotations.

\paragraph{Deployment and Scalability}  
OLIM’s components (interface and learner) are Dockerized and deployable across separate machines. The learner can be scaled independently on GPU-enabled hardware, this allows olim to be run on the cloud or with on-premise setups.

\begin{figure}[ht]
    \centering
    \includegraphics[width=0.75\textwidth]{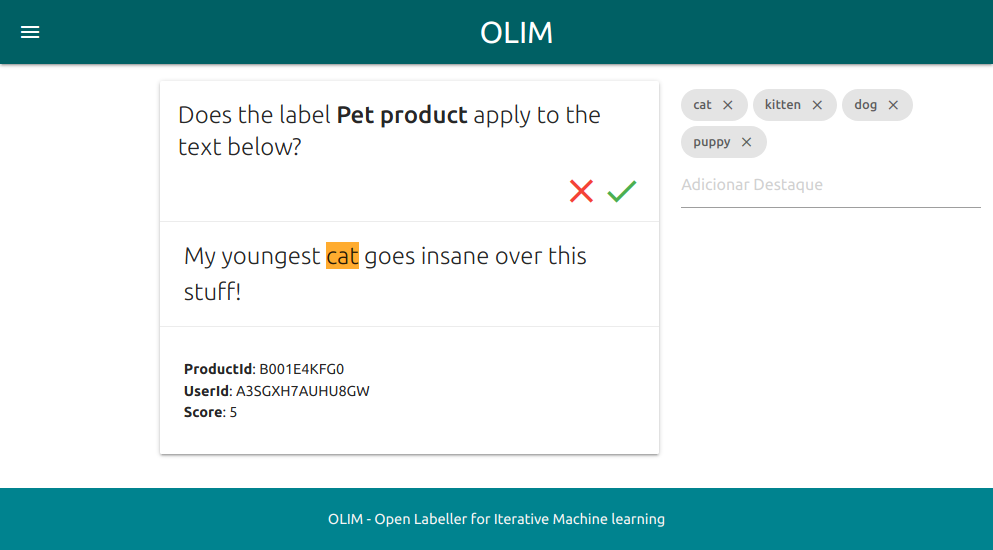}
    \caption{Interaction page for specialists. Texts are chosen to be labeled based on configurable priorities.}
    \label{fig:interaction_page}
\end{figure}

\paragraph{Interaction Page for Domain Experts (In Development)}  
A dedicated page for non-technical specialist users (e.g. medical personal) prioritizes labels requiring attention using heuristics like conflicting annotations or low model confidence. Figure~\ref{fig:interaction_page} shows the workflow. Inconsistencies can be flagged for further re-label. Those checks can ensure intra-user consistency (e.g., a specialist past decisions) and inter-user agreement.

OLIM balances scalability with usability, offering robust tools for collaborative labeling while reducing the overhead of manual labeling tasks. Its modular design supports diverse deployment scenarios, and ongoing developments focus on enhancing accessibility for domain experts through guided workflows and automated consistency checks.